\documentclass[runningheads]{llncs}
\usepackage[T1]{fontenc}
\usepackage{graphicx}
\usepackage{geometry}
\usepackage{tabularx}
\usepackage{booktabs}
\usepackage{array}
\usepackage{multicol}
\usepackage{subcaption}
\usepackage{hyperref}

\usepackage{xcolor}

\begin{document}
\title{The Dilemma of Decision-Making in the Real World: When
Robots Struggle to Make Choices Due to Situational Constraints}
\titlerunning{The Dilemma of Decision-Making in the Real World}
\author{Khairidine Benali \and Praminda Caleb-Solly}
\authorrunning{K. Benali and P. Caleb-Solly}
\institute{School of Computer Science, University of Nottingham, United Kingdom \\
\email{\{Khairidine.Benali, Praminda.Caleb-Solly\}@nottingham.ac.uk}}

\maketitle              % typeset the header of the contribution
\begin{abstract}

In order to demonstrate the limitations  of assistive robotic capabilities in noisy real-world environments, we propose a Decision-Making Scenario analysis approach that examines the challenges due to user and environmental uncertainty, and incorporates these into user studies. The scenarios highlight how personalization can be achieved through more human-robot collaboration, particularly in relation to individuals with visual, physical, cognitive, auditory impairments, clinical needs, environmental factors (noise, light levels, clutter), and daily living activities. Our goal is for this contribution to prompt reflection and  aid in the design of improved robots (embodiment, sensors, actuation, cognition) and their behavior, and we aim to introduces a groundbreaking strategy to enhance human-robot collaboration, addressing the complexities of decision-making under uncertainty through a Scenario analysis approach. By emphasizing user-centered design principles and offering actionable solutions to real-world challenges, this work aims to identify key decision-making challenges and propose potential solutions.

\end{abstract}

\section{Introduction}

In recent years, considerable progress has been made in developing robotics and autonomous systems for use within assistive scenarios \cite{Martinez}. Despite these advancements, a major challenge remains: building robots that can collaborate seamlessly with humans in an error-free, intuitive, and comprehensible manner. Achieving this requires utilizing high-level multi-modal recognition systems to efficiently manage communication and understand human requirements and goals in noisy real-world environments, which often result in ambiguity and uncertainty.

The growth of the Internet of Things (IoT) has led to a widespread presence of devices and sensors in daily life \cite{Kumar}, making them integral components of human-robot interaction (HRI). Traditional one-way interfaces focused on a single sensory modality—such as visual, tactile, auditory, or gestures—are no longer the exclusive options.

In assistive robotics, visual and auditory modes of communication are intuitive and effective \cite{Arulkumaran}, thanks to advances in modeling that enable more natural language-based interactions \cite{Li}. Nevertheless, there is still a need to improve the reliability of these interactions, which requires a high level of situational awareness. The primary objective of multimodal HRI  \cite{ANDRONAS}, \cite{Qin} in assistive scenarios is to enable individuals with accessibility needs to interact effectively with robots, thereby increasing the demand for contextual understanding to improve performance in ambiguous or uncertain environments.

Advancements in human-system communication have the potential to revolutionize human-robot interactions, leading to increased collaboration, user satisfaction, and efficiency. Enhanced communication can help robots better understand human instructions, reduce errors, and improve overall task performance and user experience. Additionally, improved communication enables robots to respond more effectively to unforeseen situations, adapt to new tasks with greater ease, and better handle unexpected changes in their environment.

In real-world HRI, robots often face challenges \cite{Brooks} where certain factors are not covered in their decision-making algorithms or training \cite{Honig}. Communication and interaction with humans can potentially improve a robot's performance in these challenging situations \cite{Zhang}, enhancing the repairability of communication failures. However, several constraints due to disabilities and impairments (visual, physical, cognitive, and auditory) are not always considered in the design of HRI hardware and software \cite{van Maris}.

Our objective is to tackle these challenges and explore potential strategies for achieving safe management of failures and learning in the context of human-robot collaboration in complex and dynamic real-world environments. This work can serve as a valuable basis for analyzing and reflecting on the highlighted challenges, planning comprehensive trials, and evaluating different scenarios in user studies. A better understanding of how interactions can be personalized for people with various disabilities and clinical requirements, as well as considering environmental factors like noise, lighting, and clutter, can inform the design of robots. This includes considerations of their physical form, sensing capabilities, mechanisms for movement, and overall behavior.

\section{Human-Robot Interaction}
Improving human-robot collaboration to resolve potential repairable failures \cite{Tolmeijer} and learn how to manage safely in a “messy” real-world environment is a challenging endeavor. The space around, and the distance between, the robot and the person can also impact the interaction, particularly when considering accessibility needs, person-environment interaction guidelines should also be considered \cite{Caleb-Solly}.  One potential avenue for progress can be achieved by designing more adaptable and customizable robots. Furthermore, providing developers and designers with a variety of benchmarks \cite{khanna},  for their methods and solutions, and to help refine and optimize their approaches.

\subsection{Optimal solutions in new scenarios}

In healthcare assistive technologies, human-robot interaction is evolving into a critical sector where the nuances of human factors require proactive consideration and prioritization to ensure effective collaboration, avoid errors, and optimize patient care \cite{Esterwood}. However,
it is evident that the majority of failure situations in the healthcare sector are caused unexpected situations for the system, in  which  small discrepancies or variations in the system's inputs or conditions can impact on the system's performance or behavior. In addition, the absence of personalized and adaptable human interaction, underscores the significant impact of human factors on resolving the consequences of failure. Consequently, it is crucial to better understand situations can result in failure, and prioritize  safety and identify mitigations to minimise risk ensure safer and resilient human-robot interaction. 
Furthermore, safe and efficient human-robot interaction is crucial in a Human-Robot Collaborative systems \cite{jsan}, particularly when designing the robot as a general assistant that offers flexibility and freedom to users. 
Given the critical role of feedback and trust within such systems, novel approaches are necessary to facilitate seamless communication between humans and robots. To realize this objective, the development of dedicated Human-Robot Interaction Modules is imperative. These modules should be designed with consideration for potential unforeseen circumstances, necessitating advanced and multimodal interfaces that cater to individuals with impairments.
Furthermore, it is crucial to comprehend the influence of failures on trust and perception, and to address their impact through strategies such as failure recovery and explainability. 

In various scenarios, robots are often designed to provide assistance with daily activities. However, in many cases, these robots are trained to make decisions based on specific scenarios and feedback from sensors. The robot may face several different types of limitations that prevent it from carrying out the next action, such as an obstacle blocking its path, noise in the environment affecting the quality of voice-based interaction or an unidentifiable target. 
Developers often rely on identifying a range of various use cases or scenarios to try and expose possible limitations or gaps in functionality in their applications. Often hazard assessment approaches, such as SHARD-UML (Software Hazard Analysis and Resolution in Design, a hazard analysis technique which is a variant of HAZOP -Hazard and Operability Analysis) and STPA (Systems-Theoretic Process Analysis) are used to analyze safety and can help to highlight what errors the robot can encounter \cite{Delgado}. STPA is a holistic, systems-based approach to investigating accident causation. It departs from traditional reliability-based models, focusing on design flaws and unsafe interactions among non-failing components. This comprehensive framework considers a broader range of variables, making it adaptable to complex technologies.
The SHARD-UML framework identifies hazards in human-robot interaction systems by applying guide words (Omission, Commission, Early, Late, Value) to UML diagrams, evaluating them based on severity levels from no hazard to high risk of user annoyance or injury.
 
In literature, several methodologies have been used to identify errors in tasks that require human-robot interaction. One such technique is Root Cause Analysis (RCA) \cite{Ji}, which identifies the underlying causes of system failures by systematically examining the sequence of events leading to the failure. Similarly, Fault Tree Analysis (FTA) \cite{Chen}, \cite{Ferguson} focuses on visualizing possible failure scenarios using a logic diagram, helping to identify potential failures and prioritize system improvements. 
Other studies have applied general techniques for identifying human errors in robotics. Human Error Analysis (HEA) investigates and categorizes errors made by humans interacting with robots to better understand failure mechanisms and develop countermeasures. Event Tree Analysis (ETA)  \cite{KHODABANDEHLOO} examines and maps out possible event chains triggered by an initial failure, enabling the identification of potential failure scenarios and their resulting consequences.
Formal methods, such as model checking \cite{Baier} and temporal logic \cite{Pnueli}, provide rigorous error detection through mathematical verification of system behaviors, but can be resource-intensive and challenging to apply to complex systems. Behavior-based methods \cite{Nocentini}, which analyze real-time human and robot behaviors using sensors and cognitive modeling, can detect deviations and predict errors based on human cognitive states. While effective in dynamic scenarios, these methods require extensive sensor infrastructure and accurate behavioral models.  
However, real-world environments can change dynamically, and not all possible scenarios can be covered in terms of decision-making algorithms. In this situation, the robot may need to interact with humans to ask for help, when the decision-making algorithm fails to find the optimal solution, especially in cases where physical obstructions are encountered, or instructions are ambiguous. By training robots through adjusting their decision-making based on human guidance, we can also improve their performance.

\subsection{Coping with changing preferences}
Humans can be unpredictable and may change their preferences, leading to inconsistent behavior in similar situations \cite{Izquierdo}. In the same scenario, an individual may alter their preferences, such as a shift in their goals for a similar situation or changes in the environment itself. In an assistive scenario, the human user might have a illness or long-term condition, causing fatigue, memory or cognitive changes, difficulty in speaking etc., due to which they might want to change how they want a task completed, or change the way that they want to interact with the robot. In such scenarios, ability for explicit or implicit communication that helps the robot adjust quickly and easily to the change can provide a user with greater control over the situation. So rather than perceiving robots as having capabilities limited to repetitive scenarios, considering means to adapt with agility to the changes is needed.

\subsection{Socio-economic situations, cultures and attitudes affecting preferences}
Robots can be trained using various datasets to make decisions in specific tasks and scenarios. However, the dataset used for training could be generated based on specific socio-economic situations, attitudes and cultures, which may result in decisions that depend on the attitudes present in the training dataset. 

One potential area where culture plays a role is cooking or meal preparation by robots. For example, preparing a cup of tea can depend on the cultural background that the designers and developers drew on during the learning phase, but it may still not be sufficient as individuals, even from the same culture, have their own preferences. Similar issues can arise from differences in responding within different social situations, where there might be concerns about how formal or informal the behaviour of robot should be, privacy concerns when providing reminders, when it is polite to interrupt and when it is not, and behaviours that might undermine the dignity of the person being supported. 

\captionsetup[table]{skip=10pt} % Example: 10pt space between caption and table
\begin{table*}[ht]
\caption{ Challenging situations for robot decision-making and the incorporation of human feedback to enhance assistance}
\centering
% \small % Adjust font size
%\begin{tabularx}{\textwidth}{|c|L|L|L|}
\begin{tabularx}{\textwidth}{|c|X|X|X|}

\hline
\textbf{Example} & \textbf{Scenario}  & \textbf{What puzzled the robot} & \textbf{Human Feedback} \\
\hline
1 & The robot struggles  in successfully delivering an item to a human as per the desired target due to a physical obstruction.  & Should I follow the same approach as in a previous similar scenario,  or should I seek assistance, or should I alter my course of action, or tell the user? &  The user suggests moving the rollator out of the way to make it easier for the robot to access the object.\\
\hline
2 & The necessity for multiple communication modalities, as some individuals may have speech impairments. 
&Should I keep asking or should I provide more options, or should I ask for help?
 
& The user is engaging with the robot through an alternative means of communication rather than verbal communication during the task.   \\
\hline

3 & The user in facing away from the robot when it arrives and doesn't respond to the robot when it speaks to it. 

& Why is the person not responding, how many times should I repeat the question or should I go away and try again?

& There is no response from the user, so the robot just goes away after a while. \\
\hline
4 & The robot was instructed to bring a pack of medicine, but instead, it found three similar-sized boxes. 

& Should I follow the same approach as
in the previous similar scenario (bring the green box). or should I seek assistance, or Should I modify my plan to take all the boxes?

& The robot takes into account the user's preferences and attempts to retrieve the correct item again. \\
\hline
\end{tabularx}
% \small % Adjust font size
\label{tab1}
\end{table*}

\section{Video exploring Some Real-World Challenges}
In order to expose some of the breakdowns in communication, we created videos (\href{https://drive.google.com/file/d/17vXESAqOMkwi9PAKMMp250ktf_-6La-O/view}{Watch the video}) that help to highlight several scenarios (Table \ref{tab1}) where a robot can face dilemmas in decision-making when it encounters difficulties in completing a task. 

In Scenario 1, a person asks the robot to deliver an item using voice and hand gestures.  The robot moves towards  the target, but struggles to access the item, which is located on the table. When encountering physical obstacles like furniture, the robot activates its LIDAR and camera sensors to detect the obstructions and recalculates its route using obstacle avoidance algorithms. If the obstruction cannot be overcome, the robot re-attempts the process to find an optimized path. However,
the robot tries out various options to resolve the issue, but it is unable to determine the most effective course of action. It is unsure whether to follow the same approach as used previously, seek assistance, or alter its course of action. The person response could generate further ambiguity, as there are two rollators present blocking its path. Furthermore the robot lacks functionality to resolve the problem, this is often the case with assistive robots designed to handle small payloads, a trade-off of being able to fit in small spaces. 

In scenario 2 the robot employs a range of communication models, including voice and vision, to engage with users. It intuitively selects the most suitable approach, adjusting its response strategy based on user feedback to facilitate effective and clear interaction. However, the scenario demonstrates the need for multimodal communication options, as the robot is unable to understand the feedback from the person. It can often be the case, if the person has a stroke, or problem with their teeth, or be drowsy, that their speech patterns change. There is also a danger of mis-hearing the person, which can result in an erroneous action on the part of the robot. Integration with other smart home or internet of things devices can be considered here. This can also improved reliability of the input in some cases. We could also consider cases where the background noise or light levels affect interaction. 

In Scenario 3, the need for advanced situational awareness is illustrated. Also, this is where person's preference, or priority of the task needs to be considered. Upon arrival, if the robot finds the user facing away and unresponsive, it initially employs verbal cues to capture their attention. Then, using camera data, it identifies the user's orientation and adjusts its approach accordingly. For example the person might not want to be disturbed when sleeping and the robot should wait until they are awake. For this, it needs additional data to find out whether the person is sleeping, and might actually accidentally wake them up by checking. On the other hand, it might be important to wake the person up, if the task is urgent, such as regular doses of medication. It could also be that the user has lost consciousness, and this brings in the need for more consideration of being able to sense breathing patterns and heart-rate using wearable sensors or advanced imaging. 

\begin{figure}[htbp]
    \centering
    \begin{subfigure}[b]{0.4\textwidth}
        \includegraphics[width=\textwidth]{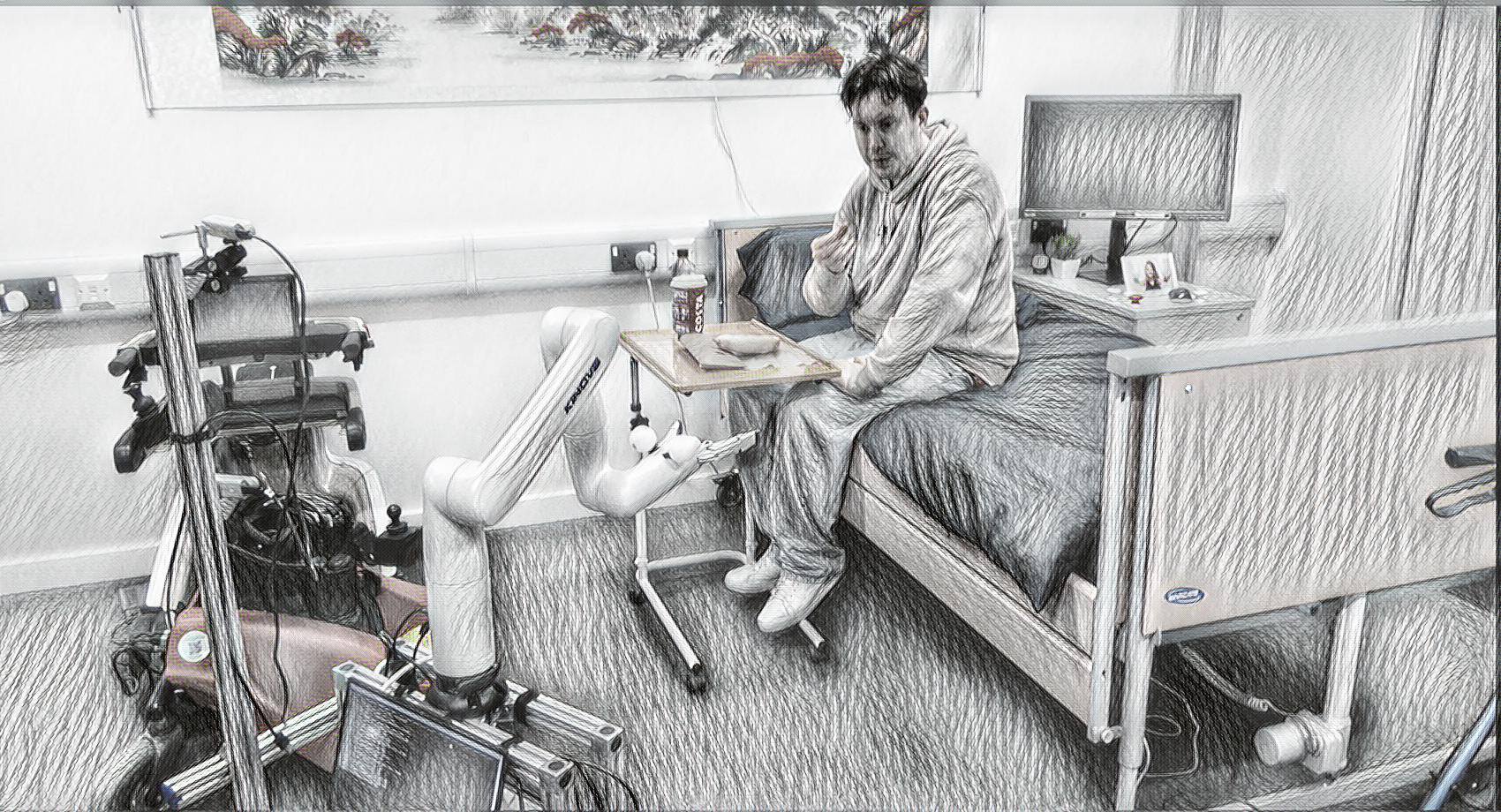}
        \caption{Person asks the robot to bring their medicine from a designated location}
    \end{subfigure}
    \hfill
    \begin{subfigure}[b]{0.4\textwidth}
        \includegraphics[width=\textwidth]{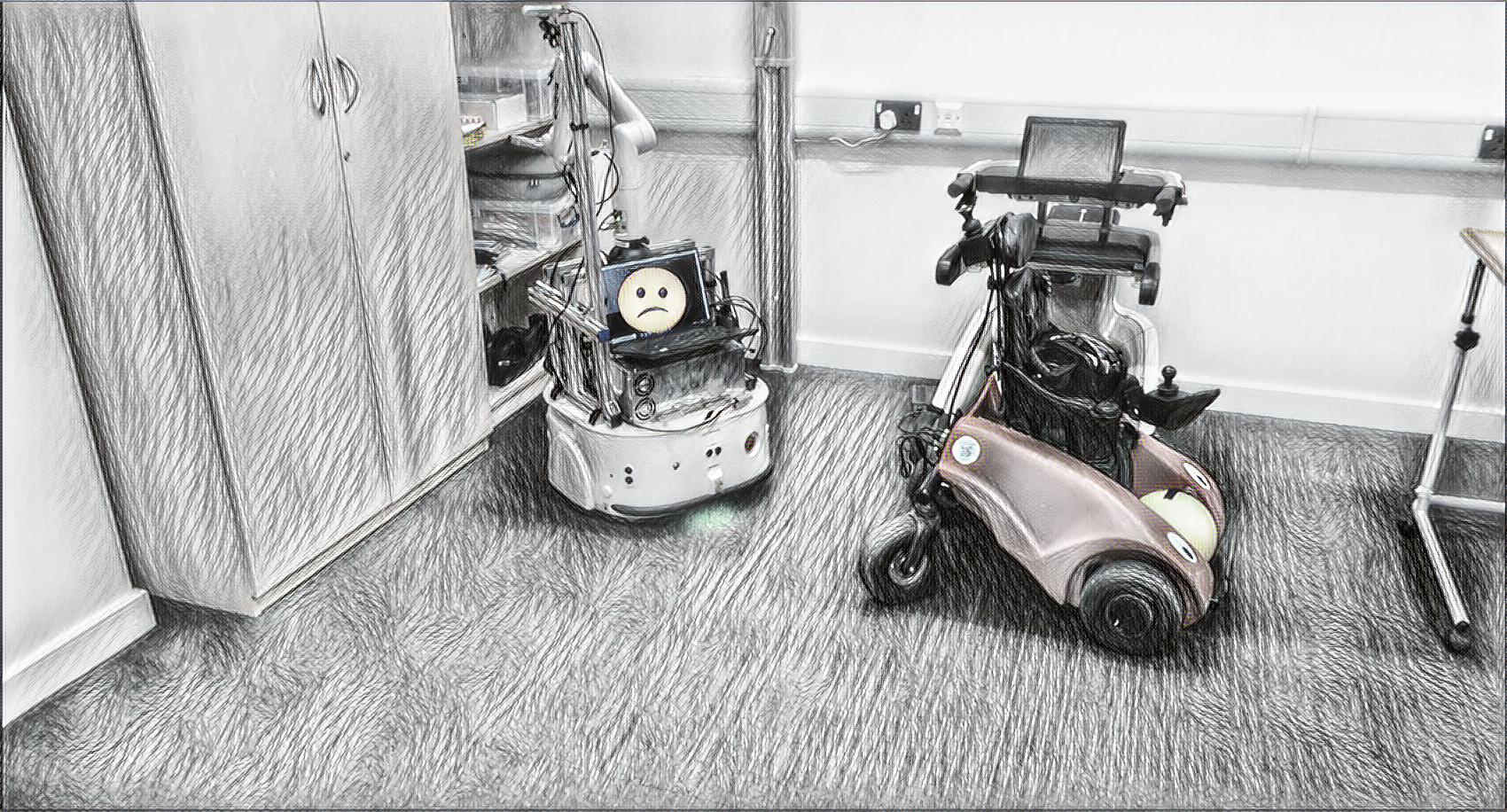}
        \caption{Robot arrives at the location but finds multiple boxes of medicine}
    \end{subfigure}
    \hfill
    \begin{subfigure}[b]{0.4\textwidth}
        \includegraphics[width=\textwidth]{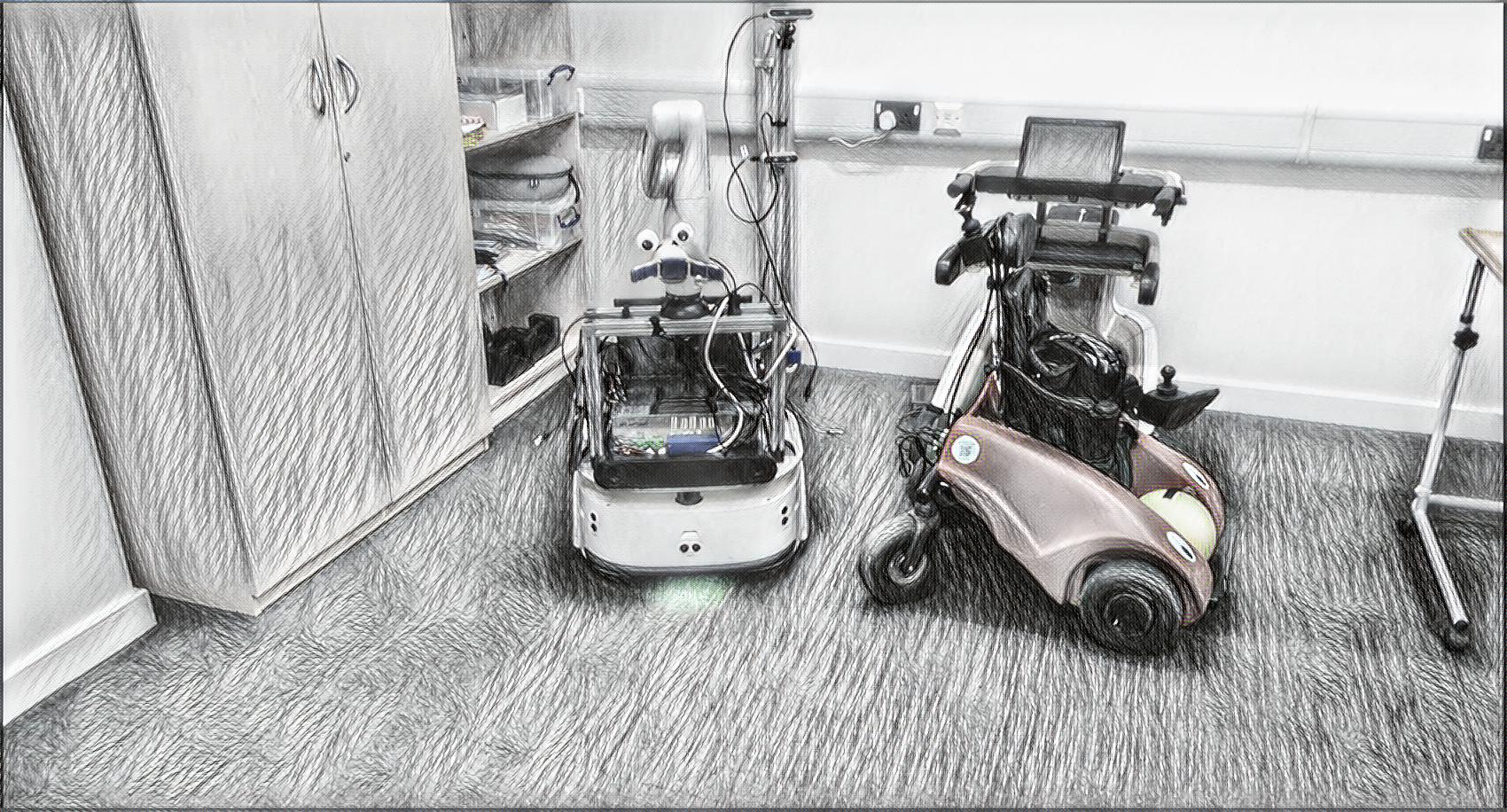}
        \caption{Robot pauses, realizing the importance of specific knowledge about the medicines for safety}
    \end{subfigure}
    
    \medskip % add some space between rows
    
    \begin{subfigure}[b]{0.4\textwidth}
        \includegraphics[width=\textwidth]{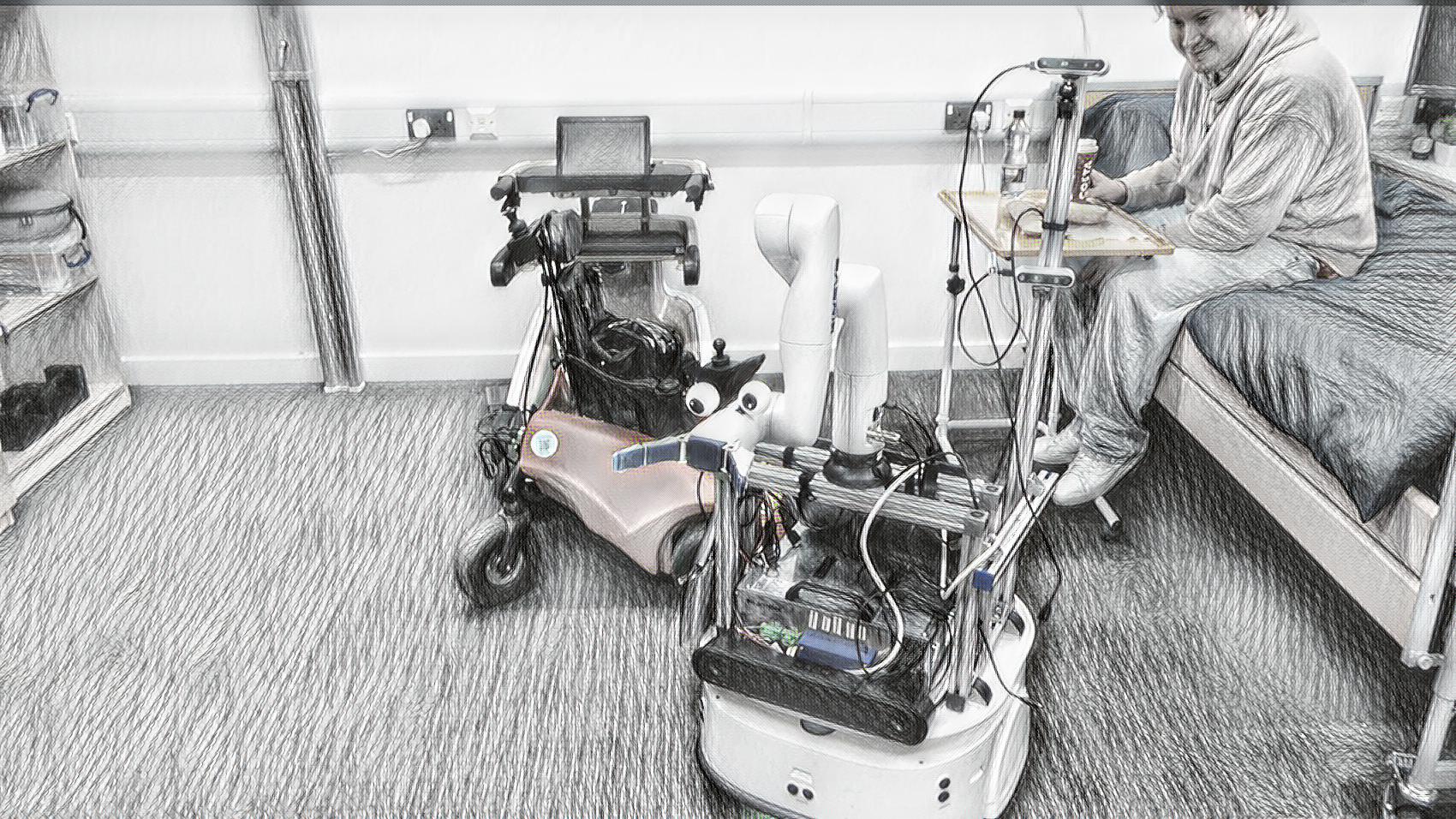}
        \caption{Robot decides to seek assistance and modify its plan to ensure the correct medicine is delivered}
    \end{subfigure}
    \hfill
    \begin{subfigure}[b]{0.4\textwidth}
        \includegraphics[width=\textwidth]{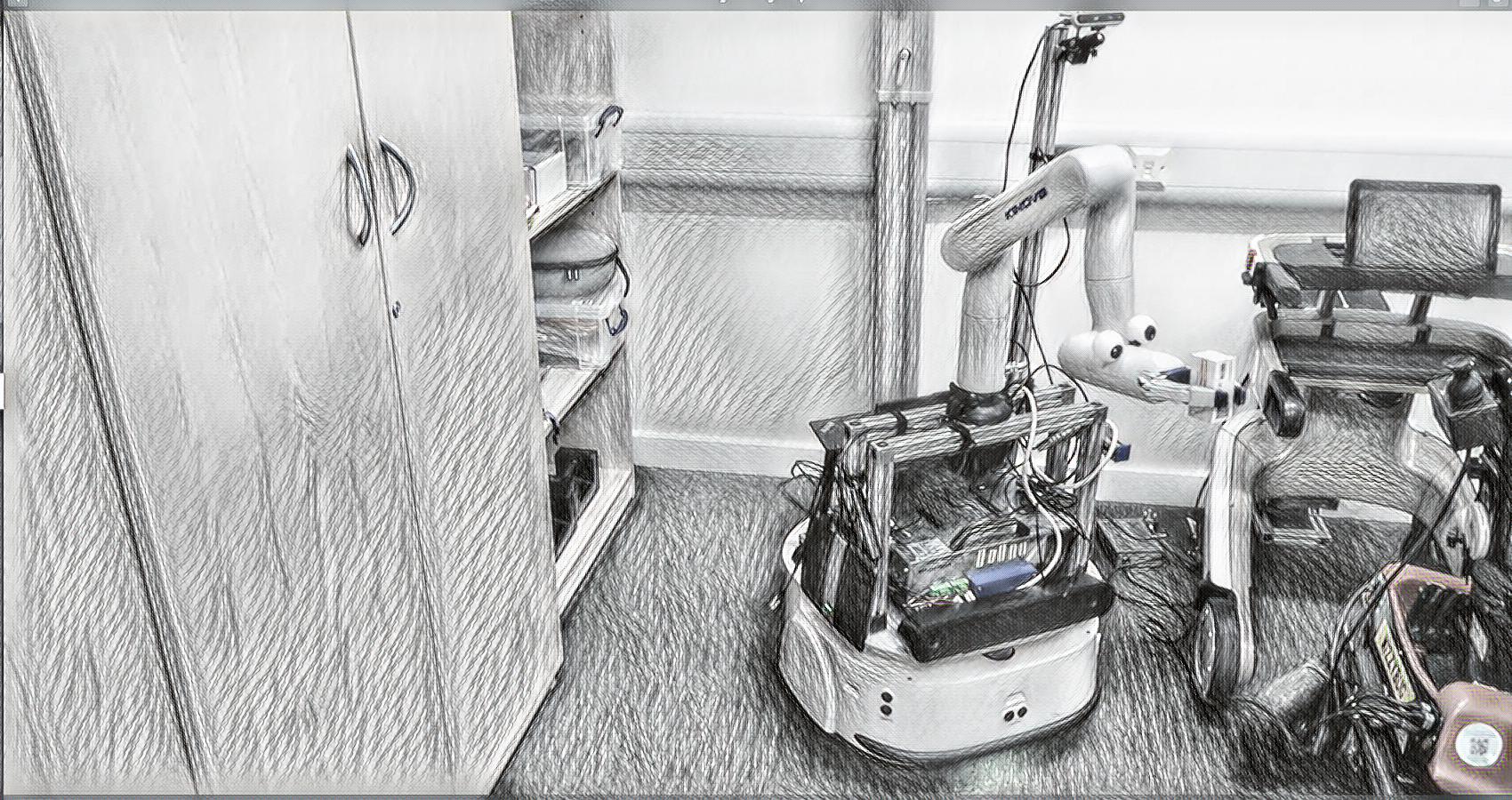}
        \caption{Robot selects the correct medication and is prepared for delivery}
    \end{subfigure}
    \hfill
    \begin{subfigure}[b]{0.4\textwidth}
        \includegraphics[width=\textwidth]{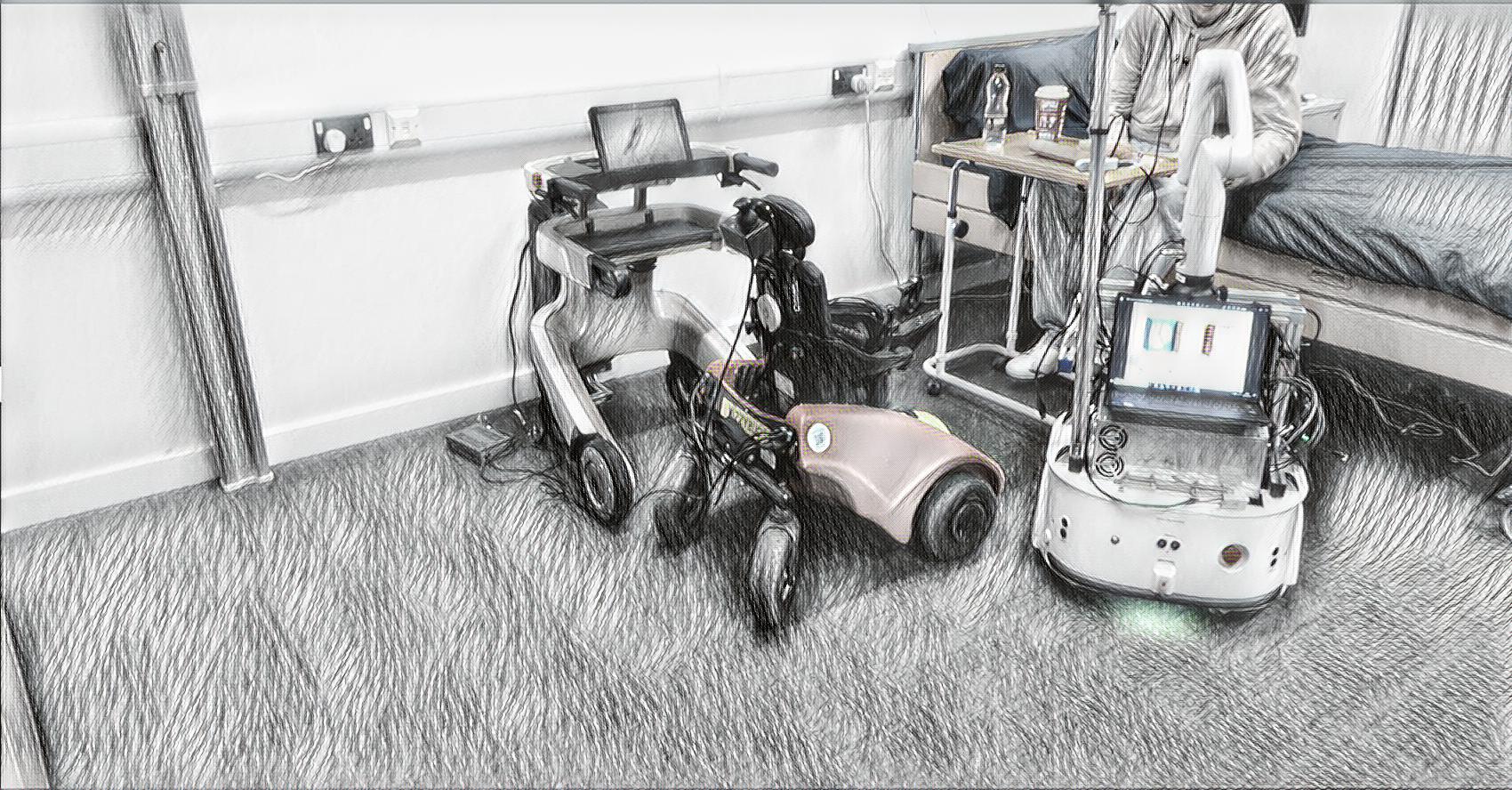}
        \caption{Robot delivers and hands over the correct medicine, completing the task}
    \end{subfigure}
    \caption{Medicine Retrieval Dilemma: Navigating Uncertainty for Safety}
    \label{fig1}
\end{figure}

In scenario 4 (Figure \ref{fig1}), the person asks the robot to bring their medicine, the robot utilizes its vision system to detect and assess distinct characteristics such as labels, shapes, or colors, enabling it to identify objects and selectively retrieve the desired one. However, the robot finds more than one box in the same place and gets confused. The robot is unsure whether to follow the same approach it might have used in a previous similar scenario, seek assistance, or modify its plan to take all the boxes. This scenario also highlights the need for including other information regarding knowledge of specific items and their use, to ensure safety of the person, based on their cognition or vision.

In all cases, the robot seeks assistance and follows the person's suggestions. The scenarios demonstrate how the optimal solution could be achieved through collaboration, which could also result in the user feeling more engaged and in control.

\section{Safety Analysis State for Robot-Assisted Patient Recovery Situation}

\begin{figure}[ht]
\centering
\includegraphics[width=5.9in,height=3.5in]{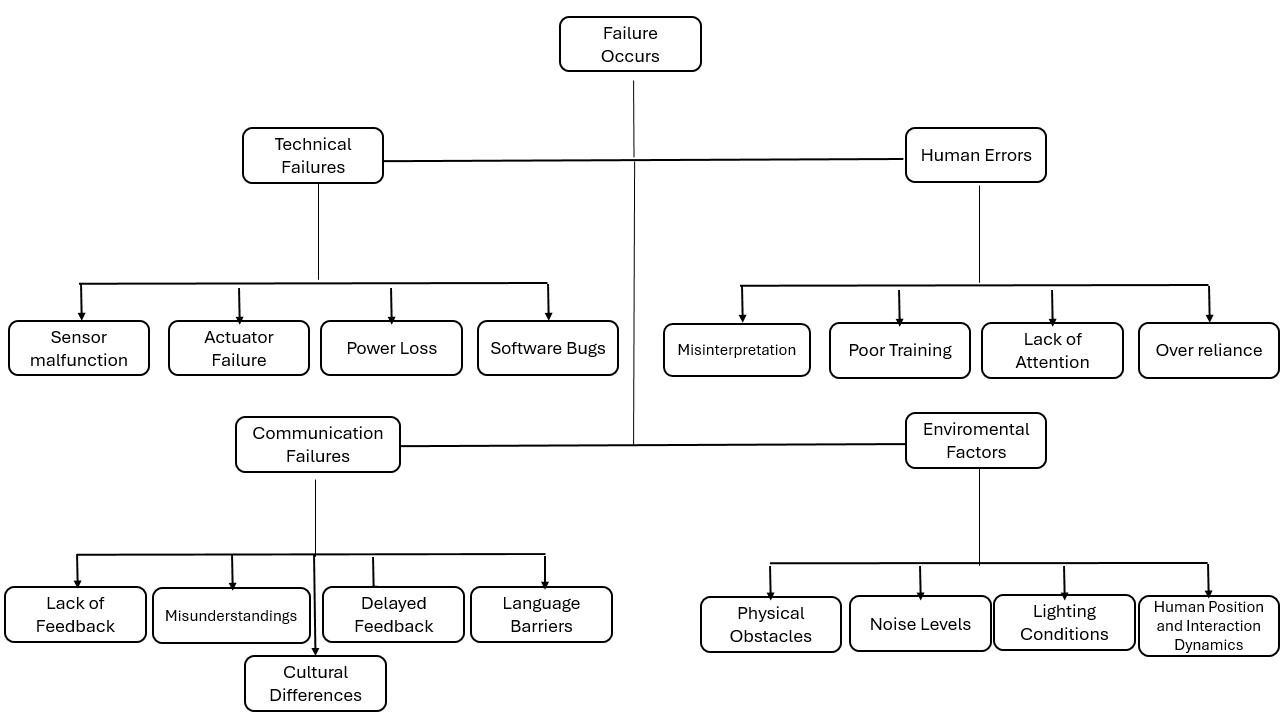}
\caption{Human-Robot Failure Taxonomy}
\label{fig2}
\end{figure}

In developing a robotic system to provide assistance for patients in the recovery period, a crucial initial step for ensuring safety involves reviewing existing methods for identifying safety concerns and potential failures during a task. For assessing the operational safety of the human-robot interaction, we utilized STPA to examine safety concerns arising from human actions, integrating insights from the Human-Robot Failure Taxonomy (Figure \ref{fig2}) to comprehensively address potential failure modes. Our selection of this technique was driven by its relevance to our specific application and its potential applicability to other similar physically assistive human-robot interaction tasks.

\subsection{STPA: User Error Analysis}

While most existing methodologies in the literature focus on identifying and anticipating component failure, STPA (Systems-Theoretic Process Analysis) takes a novel approach by adopting a holistic, systems-based framework to investigate accident causation. This departure from traditional reliability-based models allows STPA to surface a growing class of accidents caused by design flaws or unsafe interactions among non-failing components, such as those highlighted in \cite{leveson2018stpa}.

\begin{figure}[ht]
\centering
\includegraphics[width=6in,height=3.7in]{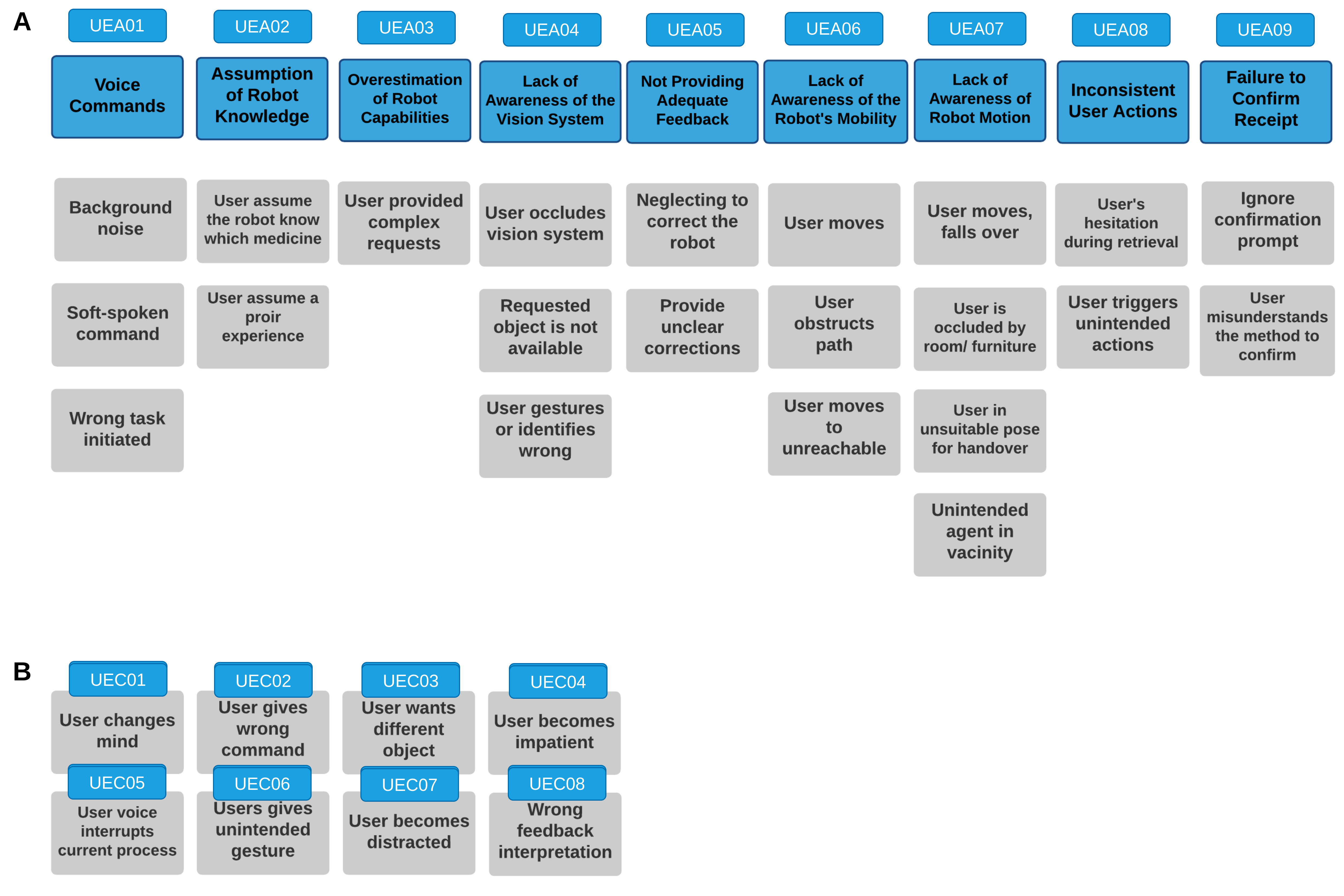}
\caption{(A)User error analysis in a medicine handover scenario 4 (B) User errors common to any task scenario element.}
\label{fig3}
\end{figure}

STPA is considered a comprehensive framework that encompasses the hazards and their causes identified by traditional methods. This approach was developed to address the growing complexity of technology. However, it is essential to recognize that STPA requires customization and completion for each specific task, which necessitates considering a broader range of variables compared to traditional methods.
Within a given context, the range of possible human actions is finite but diverse. To identify the essential actions and potentially unsafe behaviors, we have selected task scenario 4 and apply the STPA method to provide an initial abstraction

For our specific use case, we conducted a detailed analysis of user errors for Scenario 4, which is illustrated in the top row of Figure \ref{fig3}A. Our analysis focused on situations where users take actions that could lead to potential hazards. 
Every task component is assigned a unique identifier for User Error Analysis purposes, facilitating hazard identification. This aspect of hazard analysis takes on particular importance, given the potential for older adults with cognitive or physical impairments, such as memory loss or mobility issues, to interact with the system in unpredictable ways, including through unintended movements, gestures, or verbal expressions.

\section{Discussion}

This paper explores scenarios related to healthcare and assistive robots, with the goal of paving the way for additional applications and potential failures in various contexts that consider both hardware and software from the robot's perspective, as well as human error in the methodology of interaction. 

From analysing the context from a robot's perspective, various methodologies from the literature have shown promising results. One such approach is the application of Unified Modeling Language (UML) for system software analysis. UML is a widely used tool for software applications, which enables a thorough breakdown of possible outcomes by creating a diagram that defines the logical connections between blocks and the reasoning flow. This methodology can help identify user-related errors that can occur at different stages of a robotic task. 
Furthermore, the SHARD (Robot Hazard and Error Analysis) methodology can be applied to identify potential errors that the robot may encounter. This allows for a focused examination of potential failure points based on the logical sequence of actions the robot may take. By applying the SHARD technique at each node or point of the design, as defined by the UML (Unified Modeling Language), researchers can systematically analyze potential errors and develop strategies to mitigate them.

The SHARD analysis framework, integrated with Unified Modeling Language (UML), can be used to conduct a thorough examination of potential hazards in human-robot interaction systems. By applying a set of guide words, including Omission, Commission, Early, Late, and Value, to each node of the UML diagram, we can  generate a comprehensive list of  potential hazards. Each hazard is then evaluated based on its severity level, ranging from no hazard to high hazard, according to the risk of user annoyance or injury. This evaluation is aligned with established severity standards, such as the Abbreviated Injury Scale (AIS) \cite{loftis2018evolution}, to ensure consistency in assessing potential risks. The identified hazards encompass a range of issues, including communication delays, unclear sensor interpretation, user distraction, and interventions by secondary actors. While the SHARD+UML approach is effective for analyzing software flow, it is acknowledged that certain human-related aspects are not adequately addressed. Therefore, the System-Theoretic Process Analysis (STPA) methodology is explored to provide additional insights into human factors and potential hazards in human-robot interactions.

Investigating robot accidents is crucial, especially with the growing presence of social robots in various settings. As highlighted in \cite{Winfield}, robot accidents are inevitable, and a framework for thorough accident investigations is essential, similar to those used in aviation and rail industries. The proposed framework aims to collect comprehensive data, identify key facts, and conduct a structured analysis of social robot accidents. This study provides valuable insights into the challenges and processes involved in investigating robot accidents, emphasizing the importance of a proactive approach to ensure the safe deployment of social robots in society.

However, real-world settings are subject to dynamic changes, and decision-making algorithms may not encompass every conceivable scenario. Consequently, the robot might require human interaction to seek assistance when the algorithm falls short of identifying the optimal solution, particularly in instances involving physical obstacles. To advance the robotic system's ability to handle similar situations when it becomes stuck with decision-making, it is essential to consider how best to adapt the algorithms and design to suit specific applications and collect more data on multimodal detection of errors and failures in Human-Robot interactions, which will help to confirm error analysis, identify patterns, and train models to mitigate failures. 

Translating these insights into actionable guidelines for developers and practitioners involves several key steps. Future research directions could involve the development of sophisticated error detection systems that effectively integrate data from multiple sensors to accurately analyze intricate interactions, thereby enabling the detection of nuanced indications of errors of or abnormal behaviors, including visual, auditory, and tactile inputs. Enhancing multi-modal communication interfaces enables robots to engage users in a highly personalized and adaptable manner, effortlessly transitioning between spoken language, textual interactions, and gestures. Furthermore, thoughtful design incorporates visually and aurally captivating alerts to ensure user attention and understanding. By prioritizing user-centric design principles, interaction flows can be tailored to individual preferences, user feedback can be seamlessly integrated, and accessibility features can be incorporated to accommodate diverse user needs. By leveraging machine learning techniques to extract and combine features from various sources, thereby enabling the seamless integration of diverse information streams and the detection of potential indicative of errors or unusual behaviors. In addition, pattern recognition techniques can allow researchers to uncover common error patterns, which can be used to refine interaction models and train error-prevention strategies, ultimately enhancing the reliability and efficiency of human-robot interactions.  
In addition, user feedback and iteration are essential to perfecting the system, accomplished through routine user testing, adoption of an iterative development methodology, and provision of post-deployment support to resolve issues, implement new features, and ensure sustained system improvement.

\section{Conclusion}
Our video, and an example hazard analysis of a specific scenario, seek to prompt designers and developers of assistive robots to think more deeply about how improved human-robot collaboration can help to address the challenges in real-world scenarios, particularly where individuals are likely to have a range of disabilities (including visual, physical, cognitive, and auditory impairments) and clinical requirements. There are also likely to be a multitude of varying environmental factors like noise, light levels, and clutter, as well as a range of daily living activities where the user will require assistance. We hope these videos will encourage other researchers to develop other such failure scenarios and draw focus to the need for more accessible and collaborative human-robot interaction, as well as guiding user studies which incorporate situations focused on different types of failures. This will help to guide the design of robots and the underlying machine learning and AI algorithms, with more creative thinking regarding their interaction, physical form, sensors, actuators, and behavioral strategies.

\section{Acknowledgments}
This work is supported by the Metrics Project, an EU Horizon 2020 research and innovation program under grant agreement No 871252, and the EPSRC Health Technologies Network+ Emergence, EP/W000741/1. Special thanks to all contributors and collaborators for their valuable insights and support throughout the Metrics project. Thanks also to
Gabriel Leach for his help in making the video. 

% \textbf{Link to the Video} \href{https://uniofnottm-my.sharepoint.com/:v:/g/personal/praminda_caleb-solly_nottingham_ac_uk/EYSn4B1XQ_1HgU2_XRn3re0Bx3lHThbS4vYZny_J5icufQ?e=KIwYxo}{(Watch the video)}

\end{document}